\documentclass[letterpaper,10pt,conference]{ieeeconf}
\IEEEoverridecommandlockouts
\usepackage[utf8]{inputenc}
\usepackage[T1]{fontenc}
\usepackage{microtype}
\usepackage{graphicx}
\usepackage{amsmath,amssymb}
\usepackage{booktabs}
\usepackage{array}
\usepackage{float}
\usepackage{flafter}
\usepackage{algorithm}
\usepackage{algpseudocode}
\usepackage[draft,bookmarks=false]{hyperref}
\usepackage{cite}
\usepackage{xcolor}
\usepackage{tikz}

\makeatletter
\def\@IEEEauthorblockAtopspace{2ex}
\makeatother

\begin{document}
\bstctlcite{aquacap_bib_control}

\title{\LARGE \bfseries AquaCap: A Training-Free Underwater Embodied Agent\\
with Code-as-Policy}

\author{\authorblockN{\parbox{\textwidth}{\centering Xiaoshi Li$^{1,*}$, Yule Xu$^{1,*}$, Chunghiu Kong$^{1}$, Yizhou Zhou$^{2}$,\\ Yang Liu$^{2}$, Hao Yang$^{1}$, Zihao Huang$^{2,3,\dagger}$, Yunxiao Shan$^{1,4,\dagger}$}}
\authorblockA{\parbox{\textwidth}{\centering $^{1}$Sun Yat-sen University \quad $^{2}$Dalian University of Technology\\ $^{3}$BIXOCEAN.INC \quad $^{4}$Southern Marine Science and Engineering Guangdong Laboratory\\ {\small $^{*}$These authors contributed equally to this work. \quad $^{\dagger}$Corresponding authors.}}}
}

\maketitle
\thispagestyle{empty}
\pagestyle{empty}

\begin{abstract}
Recent advances in vision-language-action models have stimulated growing interest in underwater embodied intelligence. However, their reliance on large-scale interaction data limits their applicability underwater, where data collection is costly and scarce. To address this challenge, we present \textbf{AquaCap}, a training-free Code-as-Policy framework for autonomous underwater navigation and manipulation. AquaCap employs a dual-layer agent that translates task instructions and environmental observations into condition-aware plans and executable control programs. Structured perception then provides the agent with semantic, geometric, and reliability-aware observations under degraded underwater conditions. A failure-aware memory diagnoses unsuccessful actions and supports closed-loop replanning and code revision. This design enables online adaptation without task-specific training or parameter updates. AquaCap achieves a 66.43\% success rate in simulation. Real-world experiments further demonstrate autonomous grasping and object transport with an ROV, including the manipulation of targets displaced by hydrodynamic disturbances.

\end{abstract}

\begin{figure}[!t]
\centering
\includegraphics[width=\columnwidth]{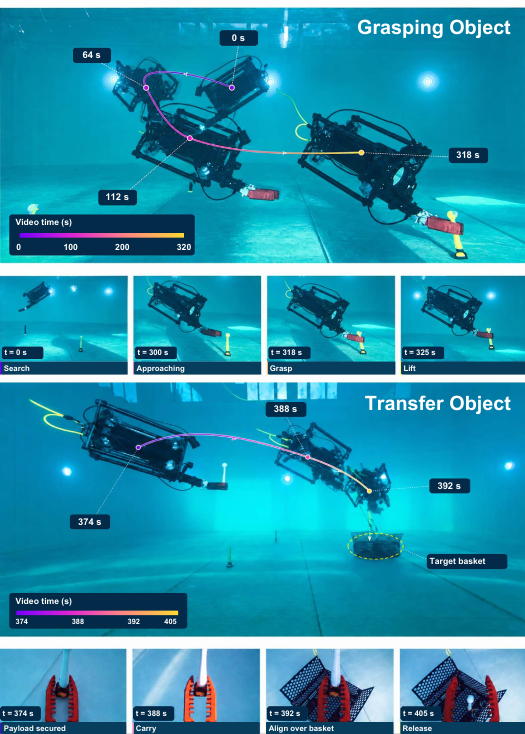}
\caption{AquaCap performing real-robot grasping (top) and transport to a basket (bottom).}
\label{fig:cover}
\end{figure}

\section{Introduction}
\label{sec:intro}

Underwater robots are becoming indispensable to the blue economy, supporting the inspection and maintenance of subsea pipelines, cables, and offshore energy infrastructure, as well as environmental monitoring, resource exploration, and deep-sea scientific research~\cite{graafland2024,nauert2023}. By extending human operational capabilities into deep, hazardous, and difficult-to-access waters, these systems reduce the risks and costs associated with conventional diver- and vessel-based operations while enabling longer-duration and wider-ranging subsea missions~\cite{huang2022,phung2024}. Consequently, improving the autonomy, adaptability, and operational capability of underwater robots is of growing importance to both marine industry and ocean science.

Underwater robotic operation remains challenging because of several fundamental characteristics of the marine environment. Complex hydrodynamics, including currents, buoyancy, and fluid resistance, make vehicle motion and control highly uncertain. Limited visibility, caused by light attenuation, scattering, color distortion, and suspended particles, reduces the reliability of visual perception. Moreover, restricted and low-bandwidth underwater communication makes continuous remote supervision and real-time data transmission difficult. 

Existing studies have addressed these challenges mainly through model-based and learning-based approaches. Classical methods combine vehicle--manipulator control, visual servoing, and teleoperation to achieve reliable underwater operation under complex hydrodynamics, limited visibility, and restricted communication~\cite{huang2022,phung2024}. However, they often require task-specific models, control strategies, and substantial human expertise. More recently, learning-based methods have attracted increasing attention. AquaBot, UMI-Underwater, and Bi-AQUA learn underwater manipulation policies from demonstrations or transferred interaction data~\cite{liu2024selfimproving,umi2026underwater,biaqua2025}. In parallel, vision-language-action (VLA) models offer a unified framework for connecting visual perception, language instructions, and robot actions~\cite{openvla2024,pi02025,gr00t2025}. 

However, existing approaches to autonomous underwater manipulation remain limited in adaptability and scalability. Conventional model-based methods encode task procedures and recovery rules in advance, making them difficult to adapt when the environment, target, or robot state changes. Learning-based methods, particularly VLA models, require large-scale interaction data that are difficult and costly to collect underwater, while simulation cannot fully reproduce real underwater conditions~\cite{openx2023,gu2026u0}. Moreover, underwater trials involve substantial deployment and recovery costs, making repeated human intervention, offline debugging, and redeployment impractical~\cite{nauert2023,phung2024}.

Motivated by the above challenges, we present \textbf{AquaCap}—combining \emph{aqua}, Latin for water, with Code-as-Policy (\emph{CaP})—a training-free alternative to data-intensive VLA models. AquaCap uses pretrained large language models to compose perception and control APIs into executable programs at runtime~\cite{liang2022code}, organized as a dual-layer planning and coding agent. A training-free perception module is proposed to transform raw sensory inputs into structured, quality-scored representations of target identity and spatial geometry, supporting reliable planning under degraded observations. To reduce reliance on human intervention and offline debugging, a failure memory further converts diagnosed failures into reusable corrections. Extensive simulation and real-world experiments validate the effectiveness of AquaCap. The specific contributions are as follows:
\begin{itemize}
\item \textbf{AquaCap-Agent.} A dual-agent Code-as-Policy architecture is introduced to separate high-level task planning from action-level code generation. Observation-based condition checks form a closed execution loop, enabling online verification, recovery, and replanning without task-specific training.

\item \textbf{AquaCap-Perception.} A structured perception method is developed to jointly represent target semantics, spatial geometry, and estimation quality, providing reliable and interpretable observations for planning and manipulation under degraded underwater sensing.

\item \textbf{AquaCap-Memory.} A failure-aware memory mechanism is developed to convert execution failures into reusable corrections. Retrieved experience guides replanning and code revision, enabling AquaCap to reflect on failures and improve subsequent actions without updating model parameters.
\end{itemize}

\section{Related Work}
\label{sec:related}
\begin{figure*}[t]
\centering
\includegraphics[width=\textwidth]{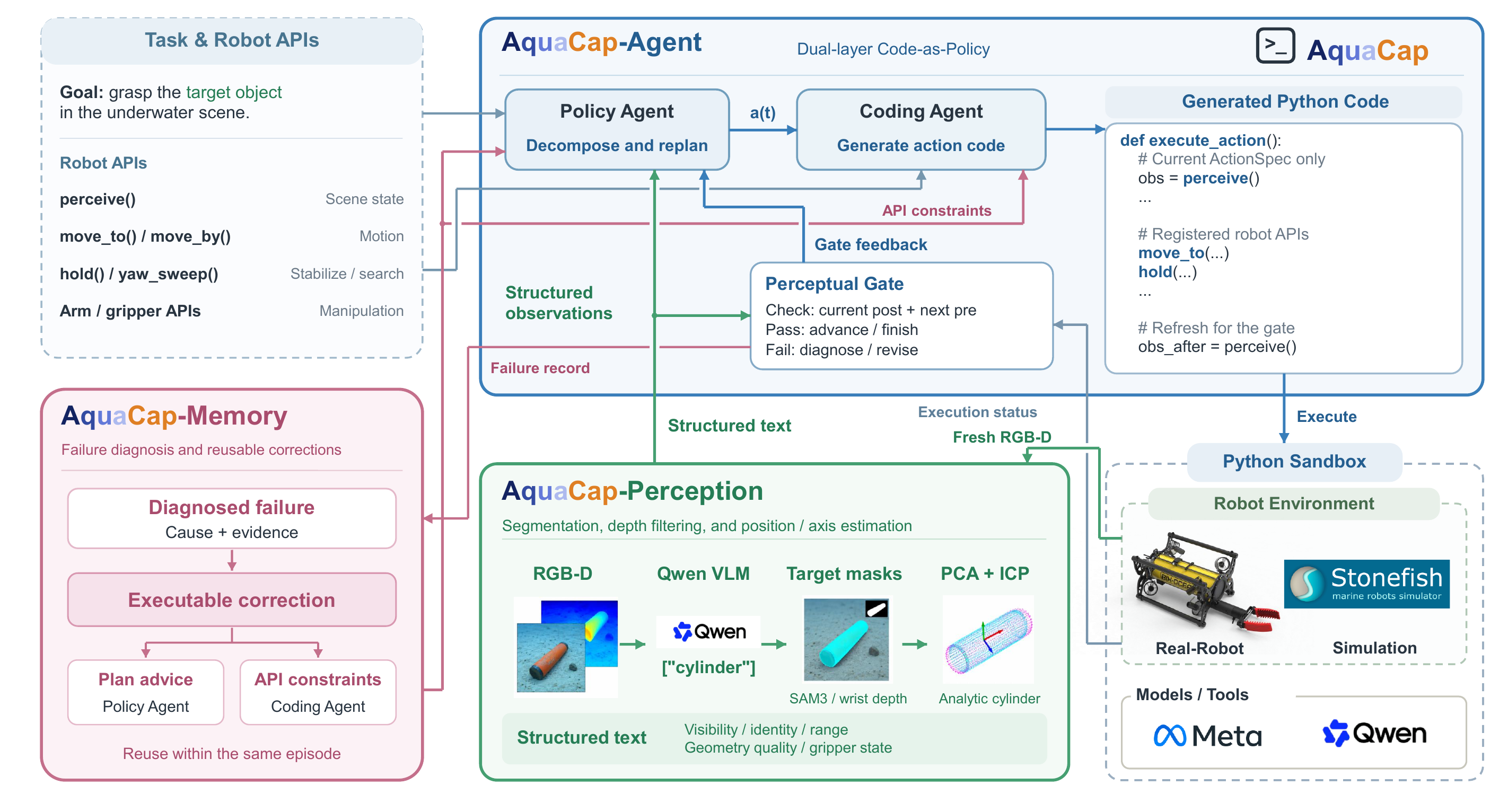}
\caption{AquaCap system overview. AquaCap-Agent handles planning and code generation, AquaCap-Perception supplies structured observations, and AquaCap-Memory reuses failure corrections.}
\label{fig:overview}
\end{figure*}

\subsection{Model-Based Control for Underwater Manipulation}

Traditional model-based methods combine vehicle--manipulator modeling, motion control, and visual servoing to perform underwater manipulation. Huang et al.~\cite{huang2022} review underwater visual servoing and the challenges posed by limited visibility, calibration errors, and model uncertainty. Shared-autonomy systems such as SHARC~\cite{phung2024} further combine engineered perception and control with operator-provided task instructions. These approaches provide accurate and reliable feedback control, but generally depend on precise models, careful calibration, and predefined task procedures. 

\subsection{Learning-Based Policies for Underwater Manipulation}
Learning-based methods acquire manipulation policies from demonstrations or interaction data.
AquaBot~\cite{liu2024selfimproving} combines imitation learning with self-learning optimization, while UMI-Underwater~\cite{umi2026underwater} and Bi-AQUA~\cite{biaqua2025} explore demonstration transfer and imitation learning under underwater sensing conditions.
More broadly, OpenVLA~\cite{openvla2024}, $\pi_{0.5}$~\cite{pi02025}, and GR00T N1~\cite{gr00t2025} use large-scale pretraining to support manipulation across tasks and environments.
Open X-Embodiment~\cite{openx2023} supports this direction by pooling demonstrations across robot platforms.
In underwater robotics, UnderwaterVLA~\cite{underwatervla2025} combines VLA reasoning with hydrodynamics-informed control for navigation.  Gu et al.~\cite{gu2026u0} introduce USIM, a simulation-based dataset for underwater robot navigation and manipulation. They also develop U0, a VLA model trained on USIM.
These advances expand policy capabilities, but adaptation remains tied to training data and its coverage of deployment conditions.

\subsection{Language-Generated Robot Programs and Execution \mbox{Feedback}}
Language-guided robotic systems build on pretrained language models to interpret instructions and coordinate task execution.
In underwater robotics, related systems support natural-language AUV piloting with hierarchical replanning~\cite{oceanchat2023,oceanplan2024}, anomaly diagnosis~\cite{aura2025}, attitude-controller adaptation~\cite{easyuuv2025}, and simulation construction~\cite{chatSim2023}.
Code as Policies (CaP)~\cite{liang2022code} provides a way to compose robot behavior without task-specific policy training by generating programs over perception and control APIs.
Related approaches ground language in robot capabilities through affordance-based skill selection~\cite{saycan2022} and language-conditioned 3D value maps~\cite{voxposer2023}.
Reliable-CaP~\cite{reliablecap2025} adds symbolic verification and interactive validation, and CaP-X~\cite{capx2026} benchmarks coding agents for manipulation and proposes CaP-Agent0, which combines multi-turn visual differencing with a synthesized skill library. However, these methods typically require large-scale training examples and interaction data to achieve reliable generalization, which are prohibitively difficult and costly to collect for underwater manipulation.

Traditional underwater manipulation methods rely heavily on task-specific designs, while learning-based approaches require large-scale interaction data that are difficult to collect underwater. Inspired by Code as Policies, AquaCap extends programmatic robot control to autonomous underwater manipulation through three key advances: a multi-agent architecture that separates high-level planning from action-level code generation; an enhanced perception module that provides quality-aware semantic and geometric estimates under degraded underwater conditions; and a failure-aware memory that transforms diagnosed failures into reusable corrections. Together, these components form a closed-loop embodied system capable of planning, execution, reflection, and self-correction without task-specific policy training.

\section{Proposed Method}
\label{sec:method}
Language-guided methods and Code-as-Policy frameworks translate task instructions into robot actions~\cite{saycan2022,liang2022code,capx2026}. A shared challenge is to keep these actions consistent with the changing physical state. Even with execution feedback, completing a command does not establish that its intended goal has been reached. Underwater, vehicle drift and target motion can invalidate intermediate results before the next action begins. Reliable execution therefore requires verifying physical outcomes and correcting failures throughout the task.

To address this challenge, we propose AquaCap, a training-free Code-as-Policy framework for underwater navigation and manipulation (Fig.~\ref{fig:overview}). AquaCap-Agent pairs a Policy agent that plans the task with a Coding agent that implements the current action. AquaCap-Perception supplies structured observations to verify action outcomes. AquaCap-Memory records failure evidence and guides subsequent corrections. Together, these components form a closed loop that advances only after the current action is verified.

\subsection{AquaCap-Agent}
\label{sec:agent}
Existing Code-as-Policy and language-guided robot systems often handle task planning and action generation within a single reasoning pipeline~\cite{liang2022code,voxposer2023}. This coupling makes planning errors difficult to separate from implementation errors, allowing a local coding failure to disrupt the entire task. Moreover, these systems often assume that generated actions are executed as intended, with limited verification of their actual physical outcomes. This assumption is particularly problematic underwater, where vehicle drift, target motion, and perception errors can cause an action to fail even when its program runs successfully. AquaCap-Agent addresses both limitations by separating task planning from action implementation: a Policy agent plans and revises the task, while a Coding agent generates code for each action. Fresh observations verify the outcome of every action and determine whether the code should be revised or the task should be replanned. Algorithm~\ref{alg:aquacap} summarizes this process.

\begin{algorithm}[!t]
\caption{AquaCap: Closed-Loop Code-as-Policy for Underwater Tasks}
\label{alg:aquacap}
\small
\begin{algorithmic}[1]
\setlength{\itemsep}{0.5pt}
\Require Instruction $l$, robot APIs, time and recovery budgets
\State $\mathcal M\gets\emptyset$; $o\gets\operatorname{Perceive}()$ \label{line:init}
\State Initialize the scene descriptor from $o$
\State $\Pi\gets\operatorname{Policy}(l,o)$ \label{line:plan}
\While{$\Pi$ has pending actions and time remains}
  \State $A\gets\operatorname{Current}(\Pi)$; $o\gets\operatorname{Perceive}()$ \label{line:current}
  \State $r\gets\text{not executed}$; check sensor validity and preconditions
  \State Stop on a sensor fault
  \If{preconditions pass}
    \State $h\gets\operatorname{Retrieve}(\mathcal M,A)$ \label{line:retrieve}
    \State $r\gets\operatorname{CodeAndRun}(A,o,h)$ \label{line:run}
    \State $o\gets\operatorname{Perceive}()$ \label{line:observe}
  \EndIf
  \State Verify $A$ using $r$, $o$, and the next action's requirements \label{line:verify}
  \State Store outcome, evidence, and any diagnosis in $\mathcal M$ \label{line:record}
  \State Stop on an unrecoverable fault
  \If{all applicable checks pass}
    \State $\Pi\gets\operatorname{Advance}(\Pi)$ \label{line:advance}
  \Else
    \State Stop if no recovery remains; otherwise use one attempt
    \State $h\gets\operatorname{Retrieve}(\mathcal M,A)$ \label{line:recover}
    \If{local repair applies and code rewrites remain}
      \State Retain $A$ and its feedback for code revision
    \Else
      \State Policy revises $\Pi$ using failure evidence and $h$ \label{line:replan}
    \EndIf
  \EndIf
\EndWhile
\State \Return plan completion or deadline expiration
\end{algorithmic}
\end{algorithm}

\begin{figure}[!t]
\centering
\includegraphics[width=\columnwidth]{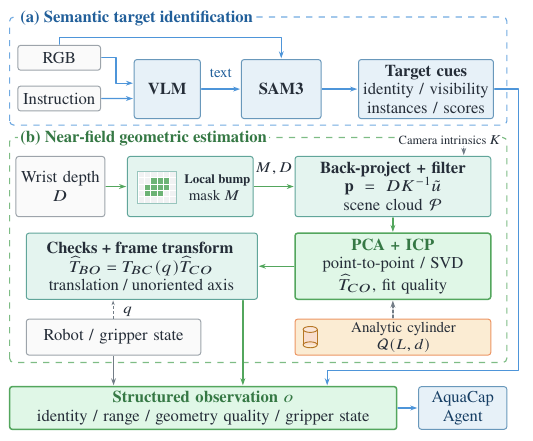}
\caption{AquaCap-Perception combines semantic masking, depth filtering, and model-based registration to produce structured geometric observations.}
\label{fig:perception}
\end{figure}

The loop begins with the task instruction $l$ and a structured observation $o$ from Perception (Alg.~\ref{alg:aquacap}, lines~\ref{line:init}--\ref{line:plan}). The observation summarizes the target and robot state from camera and robot sensor data. Policy uses these inputs to produce an ordered plan $\Pi$. Each action specifies its goal, preconditions, postconditions, and time budget. These conditions refer to observation fields and are evaluated through Boolean, numeric, or category checks. For example, an approach action can require the target to be visible before execution and within a specified range afterward.

At each iteration, the current action $A$ and an updated observation are passed to the execution loop (lines~\ref{line:current}--\ref{line:run}). Preconditions determine whether the action can start. If they pass, Coding combines the action, observation, API documentation, and retrieved corrections $h$ into a Python program. Static validation checks syntax and API use. The execution sandbox enforces motion limits, permissions, and the action's time budget. Invalid code allows a bounded number of rewrites; only validated code is executed. The function \texttt{CodeAndRun} summarizes these steps and returns their result $r$.

After execution, Perception obtains another observation to check what the action actually achieved (lines~\ref{line:observe}--\ref{line:advance}). Verification combines the execution result with sensor validity, target-estimate consistency, the action's postconditions, and the next action's preconditions. The last requirement is omitted for the final action. All applicable checks must pass before the plan advances. A program that finishes normally can therefore still fail this check. For instance, a completed alignment motion does not justify grasping if the target has moved out of position.

A failed check keeps execution at the current step and supplies evidence for correction (lines~\ref{line:record}--\ref{line:replan}). A local code repair retains the action goal and changes its implementation. Otherwise, Policy revises the remaining plan using the observed problem and Memory's guidance. A displaced target may therefore trigger localization and alignment before another grasp attempt. Unmet preconditions enter the same recovery path without executing the action. Sensor faults stop execution, and time and recovery budgets bound repeated attempts.

\begin{table*}[!t]
\vspace*{6pt}
\centering
\caption{Simulation Results}
\label{tab:main}
\small
\setlength{\tabcolsep}{5.5pt}
\begin{tabular}{@{}l cc cc cc cc c@{}}
\toprule
 & \multicolumn{2}{c}{Navigation} & \multicolumn{2}{c}{Grasping} & \multicolumn{2}{c}{Transporting} & \multicolumn{2}{c}{Tracking} & Overall \\
\cmidrule(lr){2-3} \cmidrule(lr){4-5} \cmidrule(lr){6-7} \cmidrule(lr){8-9} \cmidrule(l){10-10}
Method & $SR\uparrow$ & $SPL\uparrow$ & $SR\uparrow$ & $ASD\downarrow$ & $SR\uparrow$ & $SSR\uparrow$ & $SR\uparrow$ & $MTD\downarrow$ & $SR\uparrow$ (succ./trials) \\
\midrule
AquaCap & \textbf{86.88\%} & 0.861 & \textbf{61.25\%} & 288.20\,s & \textbf{62.50\%} & \textbf{70.00\%} & 35.00\% & \textbf{3.85\,m} & \textbf{66.43\% (465/700)} \\
U0 & 82.50\% & \textbf{0.953} & 36.25\% & \textbf{133.78\,s} & 32.50\% & 52.50\% & \textbf{55.00\%} & 4.34\,m & 47.14\% (330/700) \\
OpenVLA-ft & 31.88\% & 0.882 & 0.00\% & $-$ & 0.00\% & 0.00\% & 0.00\% & $-$ & 7.29\% (51/700) \\
CaP-Agent0 & 0.00\% & $-$ & 0.00\% & $-$ & 0.00\% & 0.00\% & 0.00\% & $-$ & 0.00\% (0/700) \\
\bottomrule
\end{tabular}
\end{table*}

\subsection{AquaCap-Perception}
\label{sec:perception}
AquaCap-Perception identifies the requested target and estimates the geometry needed to act on it (Fig.~\ref{fig:perception}). It combines semantic target identification with near-field geometric estimation and provides structured observations for execution and verification.

\textbf{(a) Semantic target identification.}
Given the instruction and an RGB image, a vision-language model (VLM) identifies the target and prompts SAM3~\cite{carion2025sam3} to segment it. The resulting masks support semantic target selection. Target visibility and confidence describe the available evidence. This stage links the language instruction to the object used for subsequent geometric estimation.

\textbf{(b) Near-field geometric estimation.}
Near the target, local surface protrusions in the wrist depth image provide a complementary mask for geometric estimation. For a selected mask $M$, valid pixels form the set $\Omega_M$. Registered depth $D$ and camera intrinsics $K$ map these pixels to a point cloud:
\begin{equation}
\mathcal P_0=\left\{D(u)K^{-1}\tilde u\mid u\in\Omega_M\right\},
\quad \tilde u=(u_x,u_y,1)^\top.
\label{eq:masked_cloud}
\end{equation}
Pixel $u=(u_x,u_y)$ has homogeneous coordinate $\tilde u$ and depth $D(u)$. Depth filtering and outlier removal yield the observed points $\mathcal P=\{\mathbf p_j\}$. Insufficient valid points produce no geometric estimate.

Geometric refinement uses a reference model built from the configured target dimensions ($L$ for length and $d$ for diameter). For cylinder-like targets, principal component analysis (PCA) estimates an initial center and axis. Iterative closest point (ICP)~\cite{besl1992method} then aligns the model points $\mathcal Q=\{\mathbf q_k\}$ with the observations. Each iteration updates nearest-point correspondences and solves
\begin{equation}
(\widehat R,\widehat{\mathbf t})=
\arg\min_{R\in\mathrm{SO}(3),\,\mathbf t\in\mathbb R^3}
\sum_{j\in\mathcal I}\|\mathbf p_j-R\mathbf q_{c(j)}-\mathbf t\|_2^2.
\label{eq:icp_objective}
\end{equation}
The correspondence $c(j)$ selects a model point for observed point $j$, and $\mathcal I$ retains accepted correspondences. Rotation $R$ and translation $\mathbf t$ define the alignment; $\|\cdot\|_2$ is the Euclidean norm. The fitted camera-frame pose is transformed to the robot base frame:
\begin{equation}
\widehat T_{BO}=T_{BC}\widehat T_{CO}.
\label{eq:frame_transform}
\end{equation}
The notation $T_{XY}$ maps coordinates from frame $Y$ to frame $X$. Frames $B$, $C$, and $O$ denote the robot base, camera, and object; hats mark estimates. Camera calibration and joint configuration $q$ provide $T_{BC}$ through arm kinematics.

Convergence, fitting quality, visible extent, mask size, and axis direction determine whether an estimate is accepted. The geometric outputs are the target position and principal axis; rotation about a symmetric cylinder's axis is unobservable. Robot primitives use accepted geometry for motion. Policy and Coding receive structured fields for target identity, range, geometric quality, camera status, and gripper state. These fields also support failure diagnosis without passing raw images to either agent.

\subsection{AquaCap-Memory}
\label{sec:memory}
AquaCap-Memory stores action outcomes and reuses corrections within the current trial, without updating model parameters. Each record links an action and scene description to its outcome and supporting evidence. Failed actions also store a diagnosed cause and proposed correction when available (Alg.~\ref{alg:aquacap}, line~\ref{line:record}). Memory $\mathcal M$ starts empty at each trial. Retrieval uses the task, current action, and an initial scene descriptor containing attributes such as target class and visibility. This descriptor is derived from the opening observation and remains fixed during the trial.

Retrieved records provide planning advice to Policy and implementation constraints to Coding (lines~\ref{line:retrieve} and~\ref{line:recover}). For example, a failed grasp can prompt Policy to repeat alignment and Coding to reduce the approach speed or adjust gripper closure. These corrections guide the next attempt; current observations still determine target geometry and verify its outcome.

\begin{table}[!t]
\centering
\caption{Simulation Task Suite}
\label{tab:setup}
\footnotesize
\setlength{\tabcolsep}{4pt}
\renewcommand{\arraystretch}{1.00}
\begin{tabular*}{\columnwidth}{@{\extracolsep{\fill}}lcc@{}}
\toprule
\textbf{Task} & \textbf{Trials} & \textbf{Time limit (s)} \\
\midrule
\multicolumn{3}{@{}l}{\textbf{Navigation}} \\
\hspace{0.6em}Goto charge station & 40 & 180 \\ %
\hspace{0.6em}Goto water tower & 40 & 180 \\ %
\hspace{0.6em}Inspect pipe (Pool) & 20 & 450 \\ %
\hspace{0.6em}Inspect pipe (Sea) & 20 & 360 \\ %
\hspace{0.6em}Scan ship (Ancient) & 20 & 270 \\ %
\hspace{0.6em}Scan ship (Modern) & 20 & 270 \\ %
\midrule
\multicolumn{3}{@{}l}{\textbf{Grasping}} \\
\hspace{0.6em}Pick red/blue/pipe0/pipe1/red$X$/blue$X$ & 480 & 600 \\
\midrule
\multicolumn{3}{@{}l}{\textbf{Transporting}} \\
\hspace{0.6em}Transfer red (Shallow) & 40 & 900 \\ %
\midrule
\multicolumn{3}{@{}l}{\textbf{Tracking}} \\
\hspace{0.6em}Follow boat & 20 & 270 \\ %
\midrule
\textbf{Total: 20 tasks} & \textbf{700} & --- \\
\bottomrule
\end{tabular*}
\par\vspace{3pt}
\parbox{\columnwidth}{\scriptsize Counts are per method.}
\end{table}

\section{Simulation}
\label{sec:sim}

\subsection{Experimental Setup and Evaluation Protocol}
\label{sec:simsetup}
We evaluate AquaCap on the 20 underwater tasks introduced by Gu et al.~\cite{gu2026u0} in the Stonefish simulator~\cite{stonefish2025}. We retain the original allocation of 700 trials per method (Table~\ref{tab:setup}). Manipulation time limits and success criteria are adapted as specified below. The simulated platform consists of a BlueROV2 equipped with a Reach Alpha~5 arm and a Robotiq Hand-E gripper. AquaCap relies only on onboard camera images and robot states, without access to ground-truth target poses. The grasping tasks marked with $X$ include same-category distractors of different colors to evaluate semantic target selection.

Policy, Coding, and VLM use DeepSeek-V4-Pro, DeepSeek-V4-Flash, and Qwen3.7-Flash through cloud APIs, respectively; SAM3 runs locally.
We fine-tune OpenVLA~\cite{openvla2024} on the USIM dataset~\cite{gu2026u0} and evaluate the resulting model as a baseline. We refer to this baseline as OpenVLA-ft. Other baselines are U0~\cite{gu2026u0} and CaP-Agent0 from CaP-X~\cite{capx2026}.
We evaluate CaP-Agent0 directly, without any adaptation to underwater tasks.
We evaluate all four methods under the same protocol with independent evaluators.
All evaluations use the same laptop (Intel Core Ultra~9 275HX, 32\,GB RAM, NVIDIA RTX~5090 Laptop GPU).

Table~\ref{tab:setup} summarizes the task counts and time limits. The main metrics are:

\begin{itemize}
\item \textbf{Success rate ($SR$).} The percentage of trials that satisfy the task-specific success criterion. Grasping requires the target to be lifted by at least 5\,cm and held for 5\,s. Transporting additionally requires the target to remain within 0.3\,m of the destination for 5\,s. Failures and timeouts are counted as unsuccessful trials.

\item \textbf{Stage success rate ($SSR$).} The percentage of transporting trials in which the grasping stage is successfully completed.

\item \textbf{Navigation and tracking.} Navigation is evaluated using success-weighted path length ($SPL$)~\cite{anderson2018evaluation}, while tracking is evaluated using mean target distance ($MTD$).

\item \textbf{Average success duration ($ASD$).} The average time from task publication to successful task completion, calculated over successful trials.
\end{itemize}

Grasping trials use predefined initial object and robot poses, with the robot facing the target. Transporting trials use fixed random seeds after all system settings are finalized. Development trials are excluded from the reported results.

\begin{table}[!b]
\centering
\caption{Grasping Ablation Results}
\label{tab:memablation}
\footnotesize
\setlength{\tabcolsep}{2pt}
\renewcommand{\arraystretch}{1.1}
\begin{tabular*}{\columnwidth}{@{\extracolsep{\fill}}lrrrr@{}}
\toprule
 & \multicolumn{4}{c}{Aggregate results (40 trials)} \\
\cmidrule(l){2-5}
Configuration & Success/$n$ & $SR$ (\%) & 95\% CI (\%) & $ASD$ (s) \\
\midrule
Full & \textbf{26/40} & \textbf{65.00} & 49.51--77.87 & \textbf{300.80} \\
Single Agent & 20/40 & 50.00 & 35.20--64.80 & 307.64 \\
w/o Memory & 21/40 & 52.50 & 37.50--67.06 & 340.89 \\
VLM-based Perception & 7/40 & 17.50 & 8.75--31.95 & 345.69 \\
\bottomrule
\end{tabular*}
\par\vspace{2pt}\parbox{\columnwidth}{\scriptsize $n$: trial count; CI: confidence interval.}
\end{table}

\subsection{Simulation Results}
\label{sec:mainresults}

AquaCap achieves an overall $SR$ of 66.43\%, outperforming U0 by 19.29 percentage points (47.14\%; Table~\ref{tab:main}). The improvement is more pronounced for grasping, where AquaCap achieves an $SR$ of 61.25\%, compared with 36.25\% for U0.

U0 learns a  policy from visual and proprioceptive inputs with target-pose supervision and supports repeated grasp attempts~\cite{gu2026u0}. Nevertheless, vehicle stabilization, target alignment, and gripper closure are generated through a learned policy, and errors arising during execution are not explicitly diagnosed and corrected. In contrast, AquaCap decomposes manipulation into verifiable stages. It uses geometric estimates to guide target alignment and evaluates the physical outcome of each stage before proceeding. When a stage fails, AquaCap uses the observed execution outcome together with corrections retrieved from Memory to revise the subsequent action or replan the remaining task. This explicit verification-and-correction loop prevents execution errors from propagating across stages.

OpenVLA-ft achieves an overall $SR$ of 7.29\% and completes no manipulation task under the tested configuration. One possible explanation is that the fine-tuning dataset lacks sufficient diversity to cover the variations encountered during testing, limiting the policy's generalization to unseen objects, poses, and environmental conditions. CaP-Agent0 also fails to complete any task in our underwater evaluation. It generates the action policy before execution and lacks intermediate outcome verification and online policy revision. Consequently, when vehicle motion or hydrodynamic disturbance displaces the target during execution, CaP-Agent0 continues to execute the previously generated action sequence based on an outdated target state. This open-loop execution prevents it from adapting to state changes and causes subsequent alignment and grasping actions to fail.

\begin{figure}[!t]
\centering
\includegraphics[width=\columnwidth]{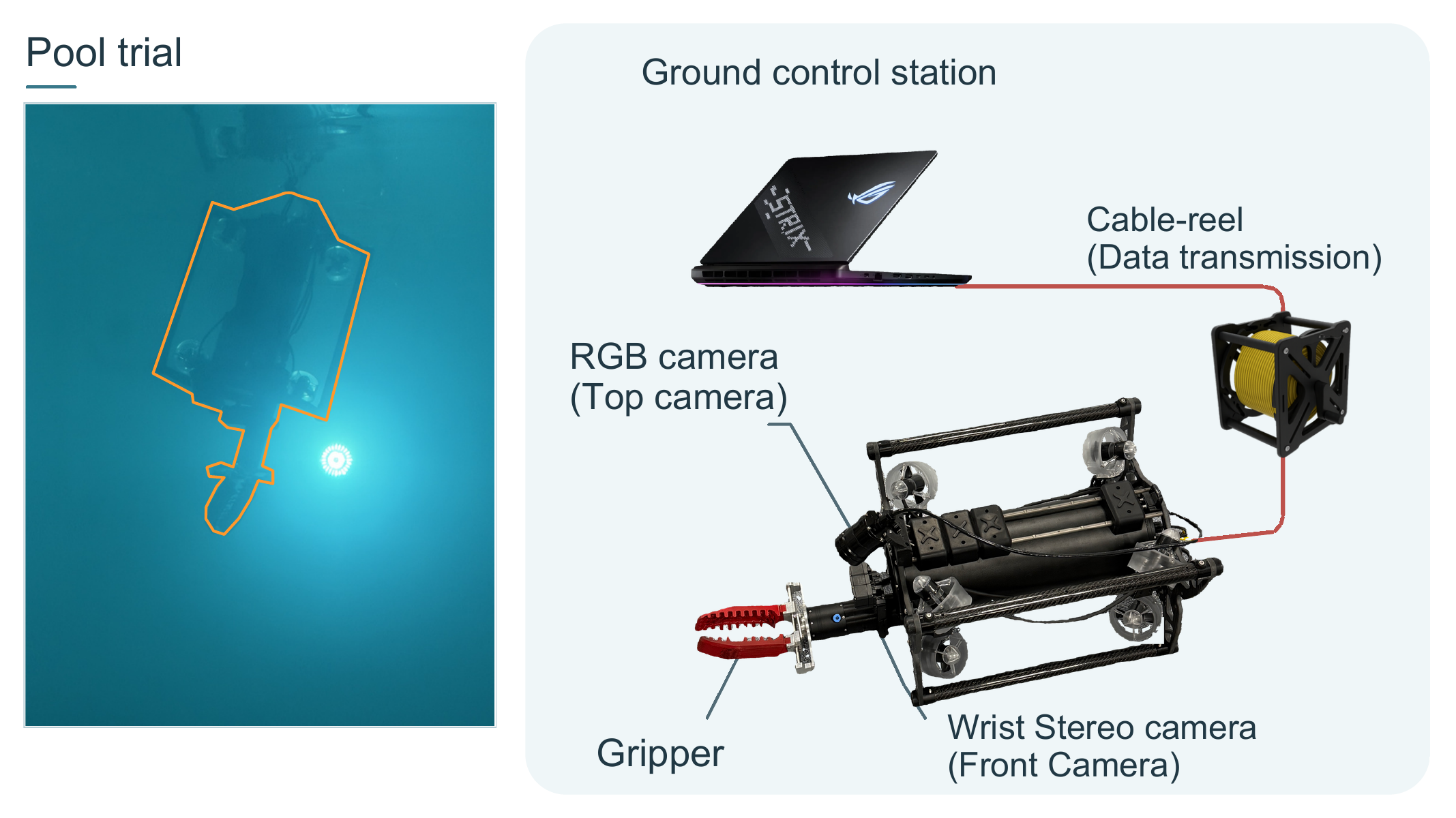}
\caption{ROV and ground-control setup used in the pool experiments.}
\label{fig:hardware_setup}
\end{figure}

\begin{figure*}[!t]
\vspace*{6pt}
\centering
\includegraphics[width=\textwidth]{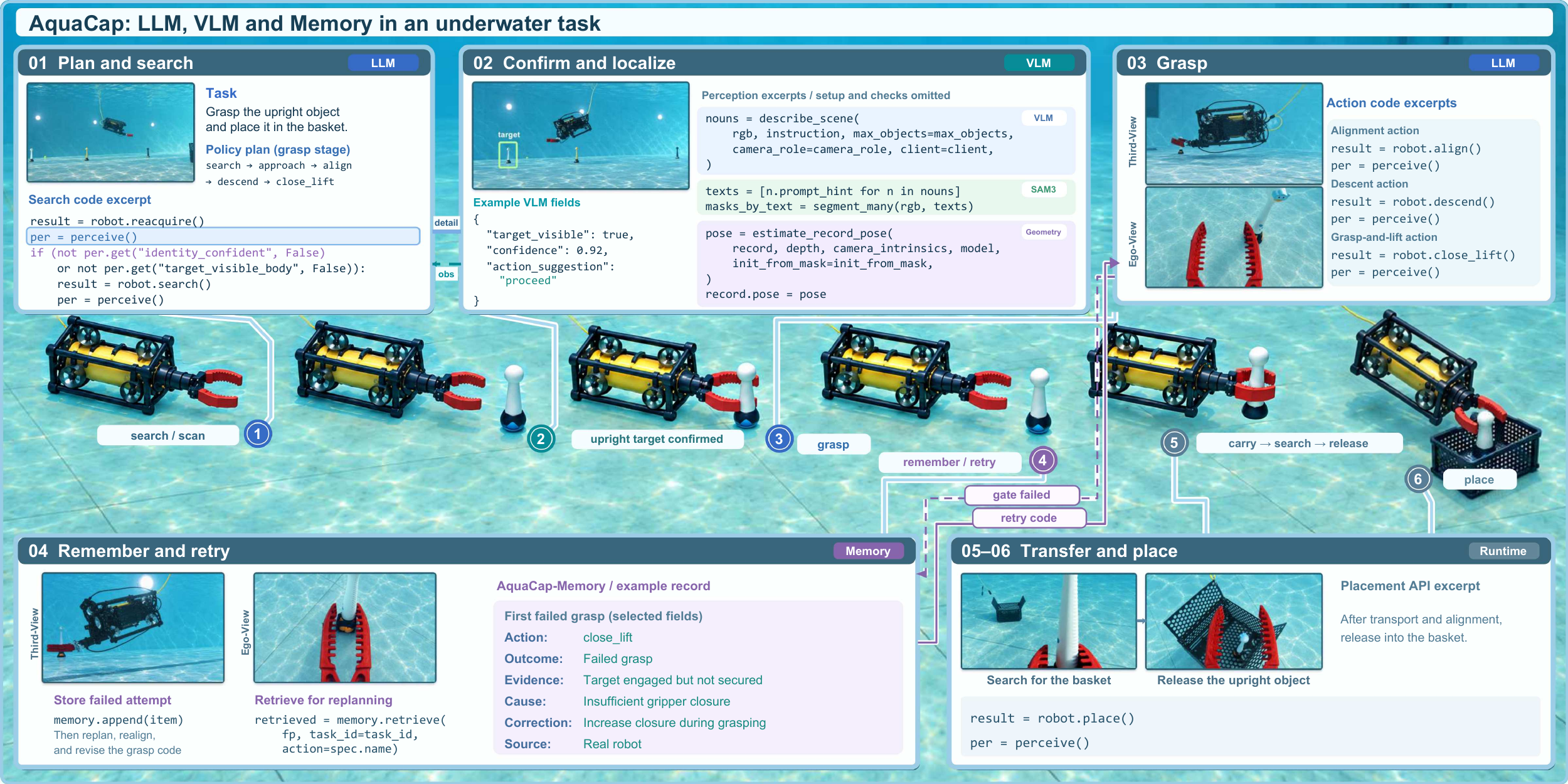}
\caption{Real-world successful execution. The ROV localizes and approaches the buoyant self-righting target, aligns the gripper, and attempts grasping. When an attempt fails, AquaCap diagnoses the cause, stores the correction in Memory, and replans using a fresh observation. After the target is secured, the robot lifts and transports it to the designated basket. }
\label{fig:real_workflow}
\end{figure*}

\subsection{Ablation Study}
\label{sec:memablation}

We conduct ablation studies on four grasping tasks: \emph{Pick red (Shallow)}, \emph{Pick blueX (Factory)}, \emph{Pick pipe0 (Shallow)}, and \emph{Pick pipe1 (Factory)}. Each configuration is evaluated over ten trials per task under matched initial scenes, robot APIs, control limits, and execution budgets. Three variants are considered: \emph{Single Agent} merges task planning and action coding into one agent; \emph{w/o Memory} disables the storage and retrieval of failure corrections while retaining immediate recovery; and \emph{VLM-based Perception} replaces our geometric perception pipeline with VLM-based target localization and depth back-projection. SAM3, depth-protrusion localization, ICP, and RGB-based metric refinement are disabled, while Agent and Memory remain unchanged. The results are shown in Tab.~\ref{tab:memablation}.

The full AquaCap achieves the highest $SR$ of 65.0\%. Merging planning and coding into a single agent reduces $SR$ to 50.0\%, suggesting that role separation in our dual-agents framework helps preserve high-level task consistency while allowing local code errors to be repaired without disrupting the overall plan. Removing Memory lowers $SR$ to 52.5\%, indicating that stored failure corrections help prevent previously diagnosed errors from recurring during subsequent execution. VLM-based Perception produces the largest degradation, reducing $SR$ to 17.5\% and failing on both pipe tasks. Grasping requires accurate metric position and orientation estimates, whereas direct VLM predictions and depth back-projection provide insufficiently stable geometry under degraded underwater observations. This result highlights geometric perception as the most influential component in the tested grasping tasks.

\section{Experiments on a Real Underwater Robot}
\label{sec:real}

\begin{table}[!t]
\centering
\caption{Real-World Evaluation Results}
\label{tab:real}
\small
\setlength{\tabcolsep}{5pt}
\renewcommand{\arraystretch}{1.08}
\begin{tabular}{lrrrr}
\toprule
Task & Success/$n$ & $SR$ (\%) & $ASD$ (s) & Replans \\
\midrule
Grasping & 2/10 & 20.0 & 391.5 & 2.5 \\
Transport & 2/10 & 20.0 & 545.0 & 3.0 \\
\bottomrule
\end{tabular}
\end{table}

\subsection{Platform and Tasks}
\label{sec:realplatform}
We deploy AquaCap on a fully vectored-thruster ROV equipped with a forward-facing RGB camera, a wrist-mounted stereo camera, and gripper position feedback (Fig.~\ref{fig:hardware_setup}). Robot commands are executed through the manufacturer's SDK, while AquaCap and SAM3 run on an external laptop with an NVIDIA RTX~5090 Laptop GPU. All trials use the complete AquaCap system without human intervention, and Memory is initialized empty at the beginning of each trial.

We conduct ten grasping trials and ten grasp-and-transport trials using a buoyant, self-righting target. Unlike a rock or another object fixed on the pool floor, the target can continuously drift and rotate under water motion, thruster wash, and manipulator disturbance, making its pose change throughout approach and grasping. AquaCap must therefore repeatedly update the target state and correct the robot motion online. A grasp is successful when the target is lifted and held for at least 5 s, while transport further requires carrying it to and releasing it into a designated basket.

\subsection{Results}
\label{sec:realresults}
Table~\ref{tab:real} reports two successful trials out of ten for both grasping and transport, corresponding to an $SR$ of 20\%. The grasping $SR$ is 41.25 percentage points lower than in simulation. The following cases examine how AquaCap responds to failures and why some corrections remain unsuccessful.

\begin{figure}[!b]
\centering
\includegraphics[width=0.90\columnwidth]{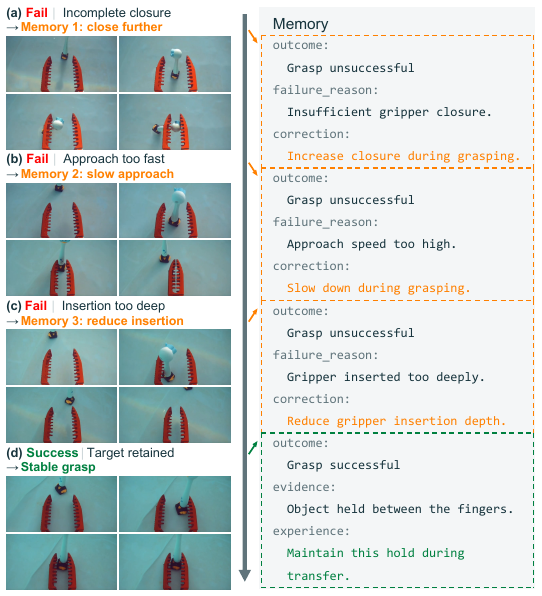}
\caption{Memory updates during the real-robot grasping sequence, expanding stage~04 (Remember and retry) of Fig.~\ref{fig:real_workflow}. Three failed attempts lead to corrections in gripper closure, approach speed, and insertion depth, followed by a successful grasp.}
\label{fig:memory_pool}
\end{figure}

\textbf{Successful trials.}
Fig.~\ref{fig:real_workflow} presents the complete real-robot workflow in a successful execution, including task planning, target search and localization, grasp execution and verification, failure recovery, and object transfer and placement. To further demonstrate AquaCap's ability to reason about execution failures and adapt its actions, Fig.~\ref{fig:memory_pool} expands stage~04 in Fig.~\ref{fig:real_workflow}, \emph{Remember and retry}. Across three failed attempts, AquaCap-Memory diagnosed \emph{insufficient gripper closure}, \emph{excessive approach speed}, and \emph{excessive insertion depth}, and generated corresponding corrections. After each failure, the Policy agent repeated localization and alignment using updated perception, while the Coding agent incorporated the retrieved correction into the next grasp program. The fourth attempt successfully secured and lifted the target, and the successful experience was stored to guide the subsequent transfer.

    \textbf{Failed trials.}
Most failures were correctly identified by AquaCap, and the proposed corrections were consistent with the observed problems. However, hardware and hydrodynamic uncertainties often caused the actual outcome to differ from the intended action. Because AquaCap controls the ROV through existing high-level APIs, it cannot directly regulate the low-level thruster dynamics. Each ROV motion therefore generates an unpredictable water flow that can substantially displace the unfixed, buoyant target, sometimes moving it outside the gripper's capture region after alignment. As a result, repeated localization, correction, and grasp attempts eventually exceeded the time limit. These findings indicate that more responsive target tracking and tighter integration with low-level vehicle control are needed to translate correct high-level corrections into reliable physical actions.

\section{Conclusion and Future Work}
\label{sec:conclusion}
AquaCap represents a step toward a more general form of underwater embodied intelligence. Rather than relying on task-specific policies, large-scale underwater training data, or fixed action sequences, it enables a robot to understand instructions, reason about tasks, generate executable actions, evaluate physical outcomes, and correct failures within a unified closed loop. The simulation and real-world results demonstrate the feasibility of this training-free paradigm for integrating underwater navigation and manipulation, and provide a foundation for underwater robots that can autonomously adapt to changing tasks and environments.

Execution efficiency, however, remains a key bottleneck. Repeated model inference, perception, and recovery improve reliability but increase task duration. Although continuing advances in foundation models are expected to reduce this cost, system-level optimization is equally important. Developing a more efficient AquaCap framework through fewer model calls, faster perception--action loops, and more effective recovery will therefore be a central direction of our future research.

\bibliographystyle{IEEEtran}
\bibliography{references}

\end{document}